\documentclass[letterpaper, 10 pt, conference]{ieeeconf}  

\IEEEoverridecommandlockouts                              

\usepackage{graphicx}
\usepackage{booktabs}
\usepackage{multirow}
\usepackage[labelfont=bf]{caption}
\usepackage{multicol}
\usepackage{dblfloatfix}
\usepackage{subfig}
\usepackage{hyperref}
\hypersetup{
    colorlinks=true,
    linkcolor=blue,
    filecolor=magenta,      
    urlcolor=cyan,
    pdftitle={Overleaf Example},
    pdfpagemode=FullScreen,
    }
\usepackage{listings}
\usepackage[font=small,labelfont=bf]{caption}

\newcommand{\shrink}{\def\baselinestretch{0.93}\large\normalsize} 
\shrink
\newcommand{\squeezeup}{\vspace*{-1.5\baselineskip}}

\title{\LARGE \bf
Skills Made to Order: Efficient Acquisition of Robot Cooking Skills Guided by Multiple Forms of Internet Data
}

\author{Mrinal Verghese$^{1}$ and Christopher Atkeson$^{1}$
\thanks{*This work was supported by the Boston Dynamics AI Institute}
\thanks{$^{1}$Mrinal Verghese and Christopher Atkeson are with the Robotics Institute at Carnegie Mellon University
        {\tt\small mverghes, cga @andrew.cmu.edu}}%
}

\begin{document}

\maketitle
\thispagestyle{empty}
\pagestyle{empty}

\begin{abstract}

This study explores the utility of various internet data sources to select among a set of template robot behaviors to perform skills. Learning contact-rich skills involving tool use from internet data sources has typically been challenging due to the lack of physical information such as contact existence, location, areas, and force in this data. Prior works have generally used internet data and foundation models trained on this data to generate low-level robot behavior. We hypothesize that these data and models may be better suited to selecting among a set of basic robot behaviors to perform these contact-rich skills. We explore three methods of template selection: querying large language models, comparing video of robot execution to retrieved human video using features from a pretrained video encoder common in prior work, and performing the same comparison using features from an optic flow encoder trained on internet data. Our results show that LLMs are surprisingly capable template selectors despite their lack of visual information, optical flow encoding significantly outperforms video encoders trained with an order of magnitude more data, and important synergies exist between various forms of internet data for template selection. By exploiting these synergies, we create a template selector using multiple forms of internet data that achieves a 79\% success rate on a set of 16 different cooking skills involving tool-use.

\end{abstract}


\section{Introduction}

Robot skills requiring the use of tools,  like cutting, peeling, scrubbing, or stirring, are particularly challenging to learn from internet data such as text, images, or videos as this data lacks physical information such as force or contact. As a result, methods that learn robot skills using internet data must integrate either other physical information like additional robot data \cite{brohanRT2VisionLanguageActionModels2023}\cite{bahlAffordancesHumanVideos2023} or use explicit methods to execute learned position trajectories on a robot \cite{bharadhwajTrack2ActPredictingPoint2024}. In addition, these methods often focus on learning relatively simpler skills, like pick-and-place skills or skills to manipulate rigid objects with articulated joints, like doors or drawers~\cite{bahlHumanRobotImitationWild2022}\cite{bharadhwajGeneralizableZeroShotManipulation2023}. To expand the range of skills we can learn from internet data, what if we endowed a robot with a large library of primitive contact behaviors, also called templates, and used internet data to select among them to execute contact-rich tool-use skills? Such a library could facilitate learning more complex skills by selecting and composing behaviors already in the library. Our primary insight is that while directly optimizing a policy for these contact-rich skills from internet data can be challenging, internet data and foundation models are well suited to select among an existing set of templates. Furthermore, we believe different internet resources, such as language and vision foundation models, can be synergistic. 

Given this paradigm of using internet data to select among a set of behavior templates, the next natural question is what forms of data we should leverage and how we should use them most effectively. In this study, we consider two popular forms of internet data for performing template selection: text and human video. To select templates with text, we use the common-sense reasoning capabilities of Large Language Models (LLMs) to consider text descriptions of templates and their applicability for a given skill. To select templates with human video, we examine two representations to compare videos of robot skill execution to videos of expert human demonstrations: a standard method from prior work using video encoders pretrained on large human video datasets and a novel method based on encoding dense optic flow.
We evaluate these approaches across eight different cooking skills with 16 total variations, including cutting, peeling, scrubbing, wiping, stirring, spreading, slicing, and scraping. Our evaluation is done with a real robot using off-the-shelf tools like spatulas, knives, peelers, sponges, and real ingredients like vegetables, sauces, and bread. We evaluate the success of different sources of internet supervision via explicit criteria and independent human evaluations. Our contributions are as follows:
\begin{itemize}
    \item We develop a paradigm for robot learning from internet data that involves selecting from a library of behavior templates.
    \item We examine the relative performance of language and human video to select among robot behavior templates to perform cooking skills and identify synergies between these two approaches.
    \item We identify a novel representation for comparing videos of robot skill execution to videos of human experts completing that skill for template selection, which increases the success rate of skill completion by 24\%
    \item Using our insights from this work, we demonstrate a robot system capable of executing cooking skills with a 79\% success rate.
\end{itemize}

 \begin{figure*}[!t] 
    \centering
  \subfloat{%
       \includegraphics[width=0.2\linewidth,bb=0 0 640 480,trim={3cm 5cm 3cm 1cm},clip]{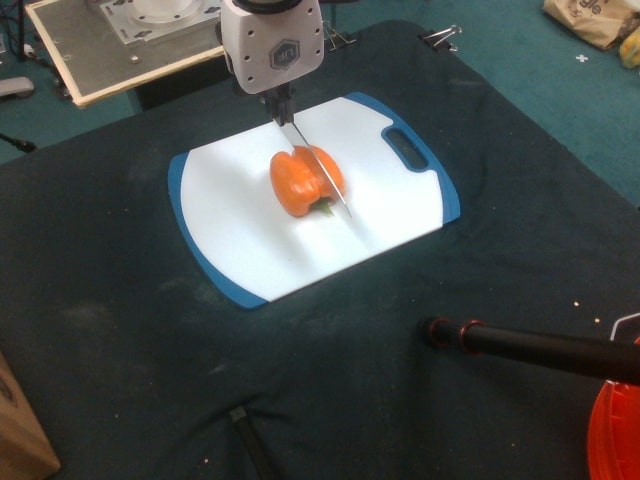}}
  \subfloat{%
        \includegraphics[width=0.2\linewidth,bb=0 0 640 480,trim={3cm 5cm 3cm 1cm},clip]{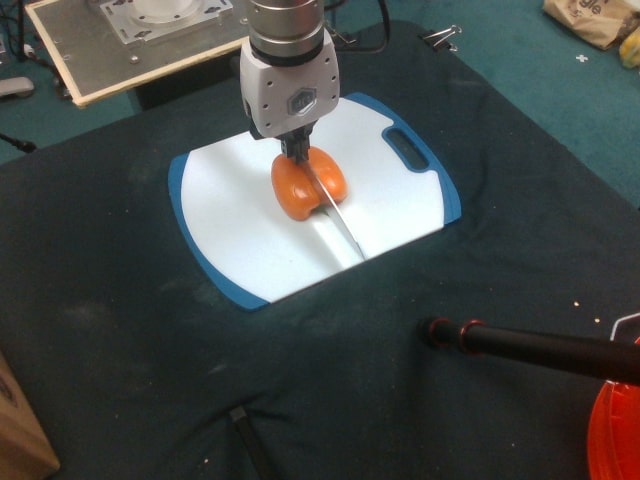}}
  \subfloat{%
        \includegraphics[width=0.2\linewidth,bb=0 0 640 480,trim={3cm 5cm 3cm 1cm},clip]{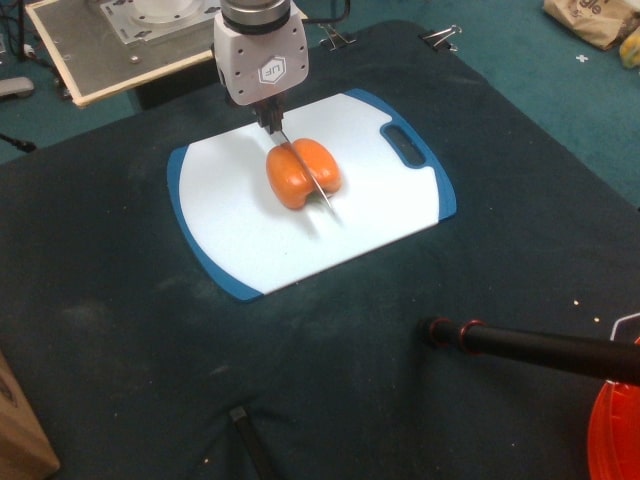}}
  \subfloat{%
        \includegraphics[width=0.2\linewidth,bb=0 0 640 480,trim={3cm 5cm 3cm 1cm},clip]{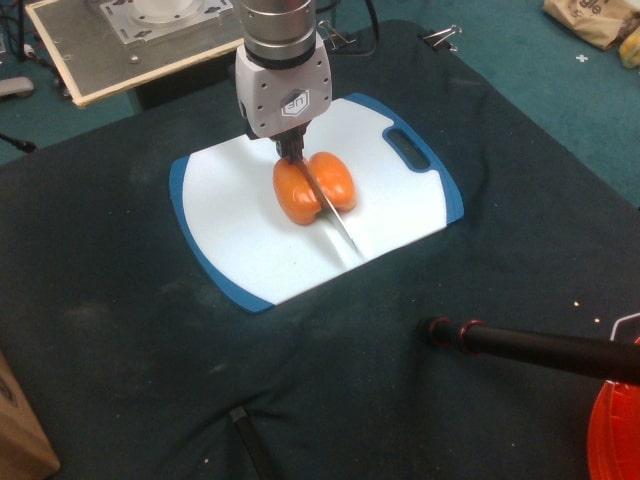}}
  \subfloat{%
        \includegraphics[width=0.2\linewidth,bb=0 0 640 480,trim={3cm 5cm 3cm 1cm},clip]{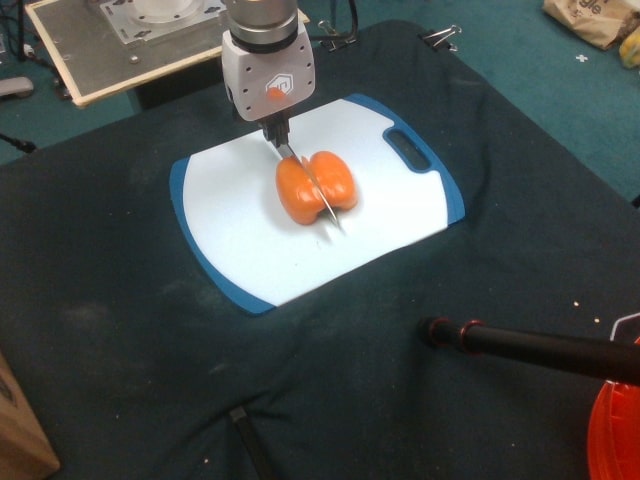}}\\
        \vspace{-4mm}
  \subfloat{%
       \includegraphics[width=0.2\linewidth,bb=0 0 640 480,trim={3cm 5cm 3cm 1cm},clip]{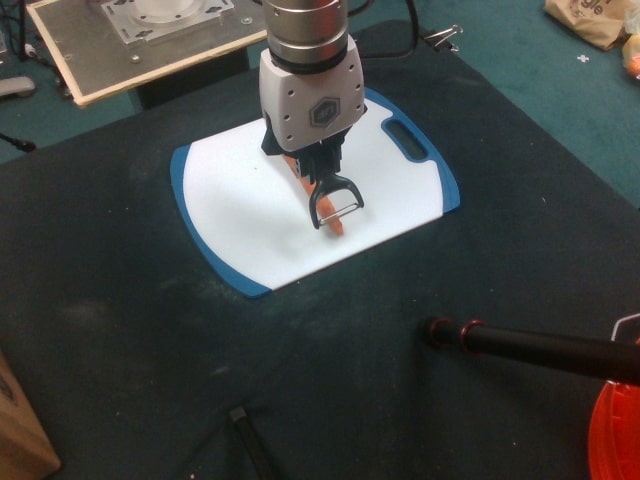}}
  \subfloat{%
        \includegraphics[width=0.2\linewidth,bb=0 0 640 480,trim={3cm 5cm 3cm 1cm},clip]{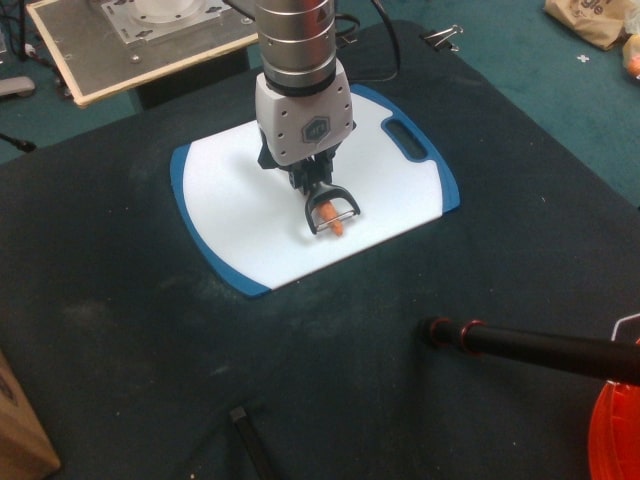}}
  \subfloat{%
        \includegraphics[width=0.2\linewidth,bb=0 0 640 480,trim={3cm 5cm 3cm 1cm},clip]{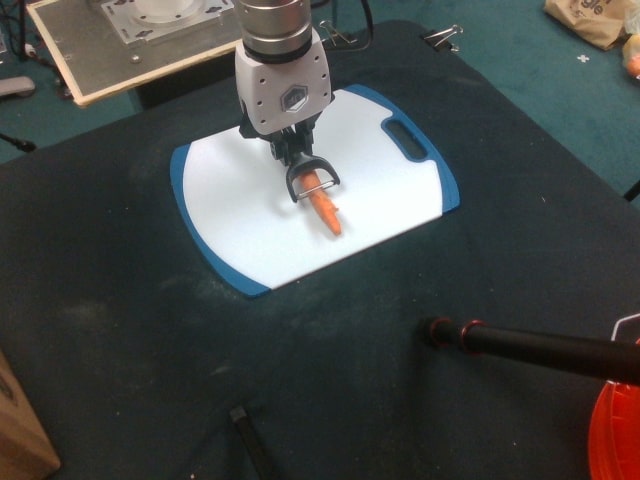}}
  \subfloat{%
        \includegraphics[width=0.2\linewidth,bb=0 0 640 480,trim={3cm 5cm 3cm 1cm},clip]{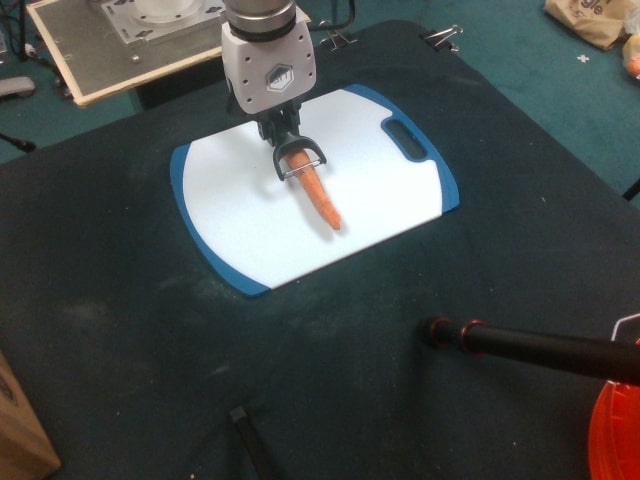}}
   \subfloat{%
        \includegraphics[width=0.2\linewidth,bb=0 0 640 480,trim={3cm 5cm 3cm 1cm},clip]{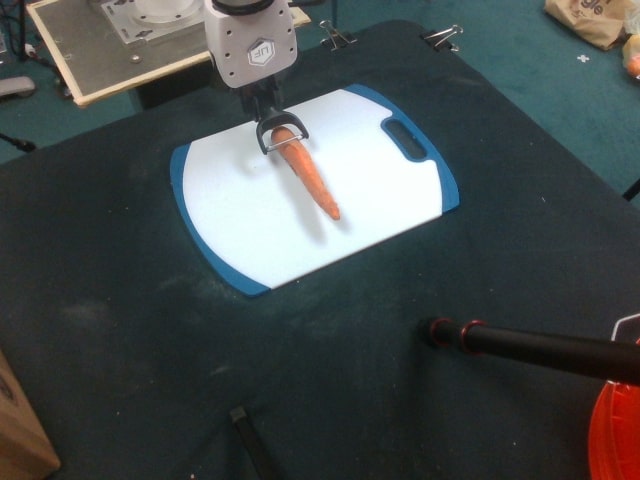}}\\
        \vspace{-4mm}
  \subfloat{%
       \includegraphics[width=0.2\linewidth,bb=0 0 640 480,trim={3cm 5cm 3cm 1cm},clip]{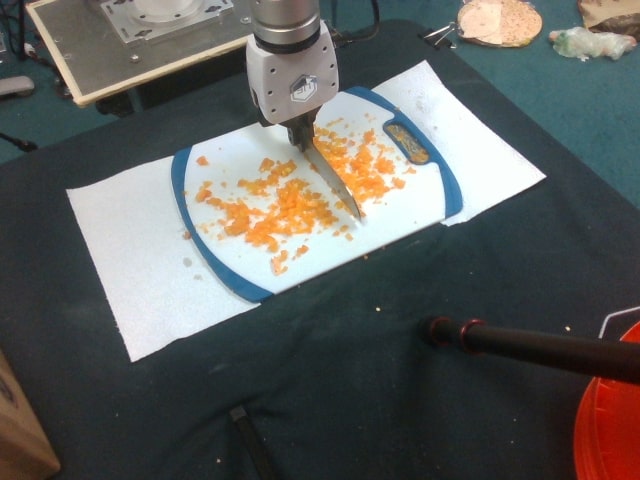}}
  \subfloat{%
        \includegraphics[width=0.2\linewidth,bb=0 0 640 480,trim={3cm 5cm 3cm 1cm},clip]{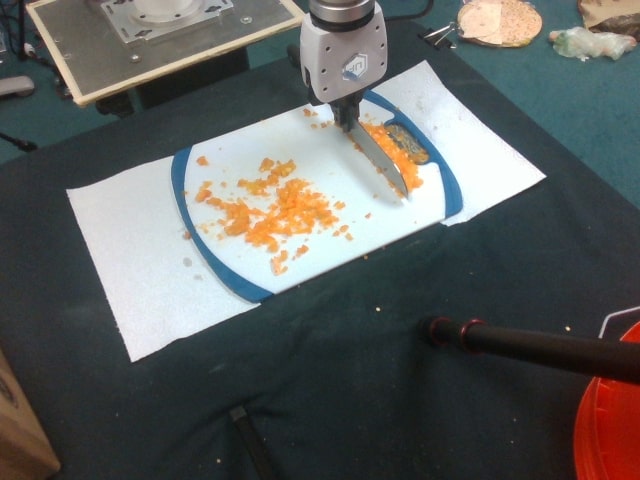}}
  \subfloat{%
        \includegraphics[width=0.2\linewidth,bb=0 0 640 480,trim={3cm 5cm 3cm 1cm},clip]{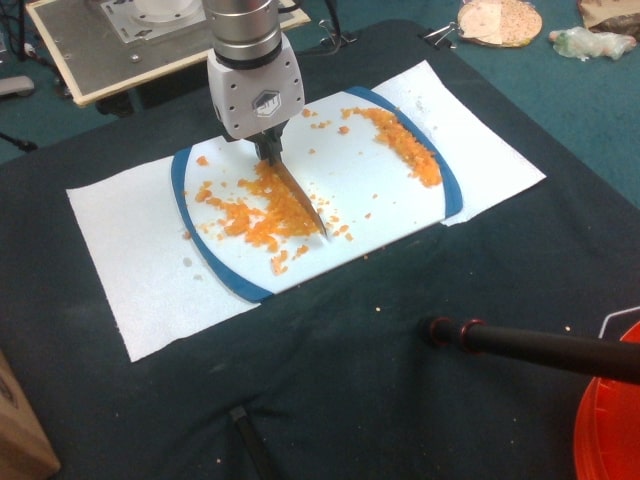}}
  \subfloat{%
        \includegraphics[width=0.2\linewidth,bb=0 0 640 480,trim={3cm 5cm 3cm 1cm},clip]{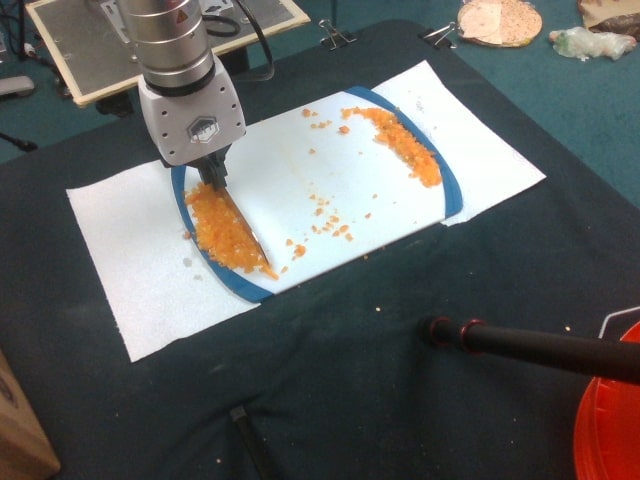}}
   \subfloat{%
        \includegraphics[width=0.2\linewidth,bb=0 0 640 480,trim={3cm 5cm 3cm 1cm},clip]{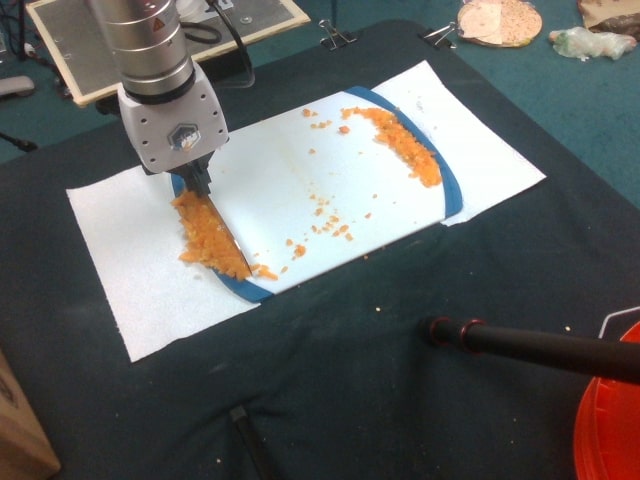}}\\
        \vspace{-4mm}
  \subfloat{%
       \includegraphics[width=0.2\linewidth,bb=0 0 640 480,trim={3cm 5cm 3cm 1cm},clip]{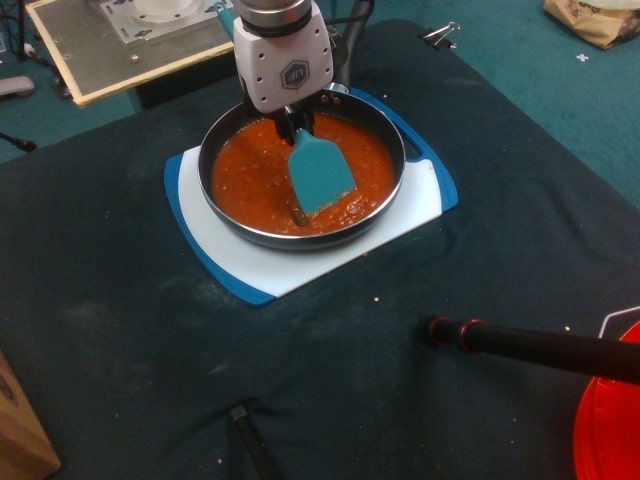}}
  \subfloat{%
        \includegraphics[width=0.2\linewidth,bb=0 0 640 480,trim={3cm 5cm 3cm 1cm},clip]{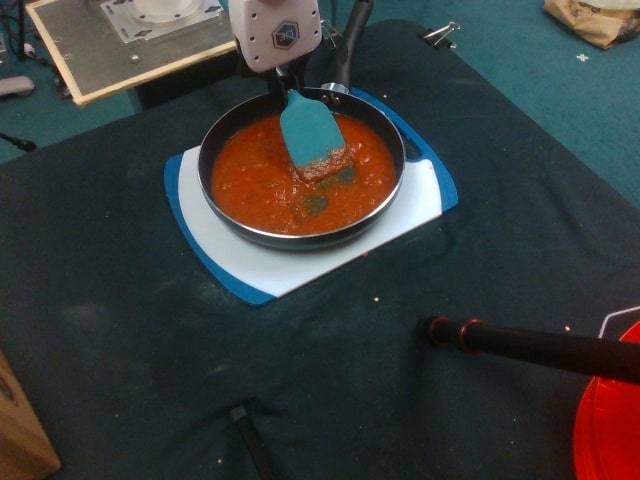}}
  \subfloat{%
        \includegraphics[width=0.2\linewidth,bb=0 0 640 480,trim={3cm 5cm 3cm 1cm},clip]{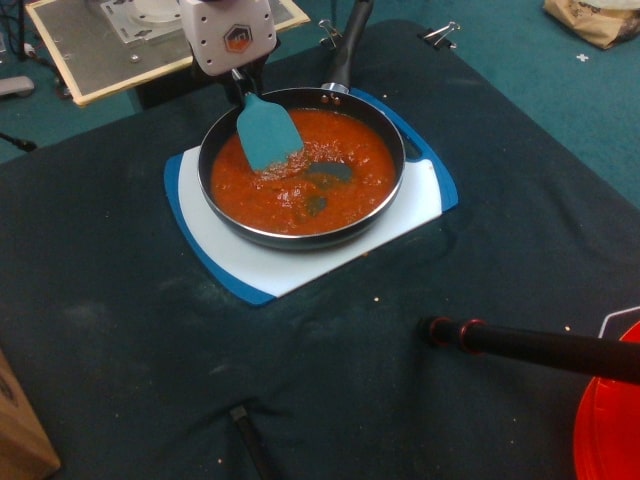}}
  \subfloat{%
        \includegraphics[width=0.2\linewidth,bb=0 0 640 480,trim={3cm 5cm 3cm 1cm},clip]{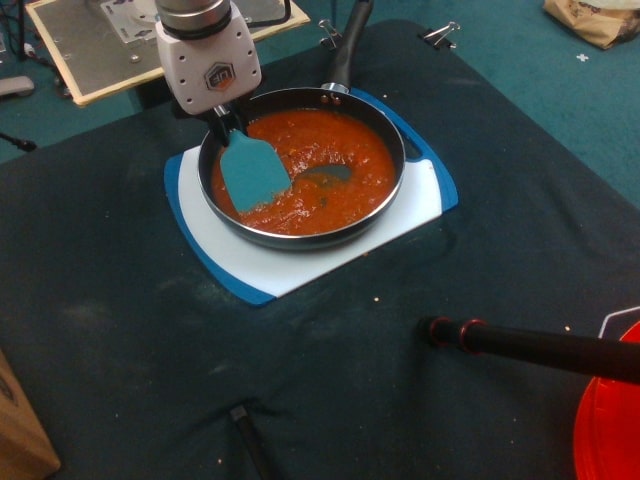}}
   \subfloat{%
        \includegraphics[width=0.2\linewidth,bb=0 0 640 480,trim={3cm 5cm 3cm 1cm},clip]{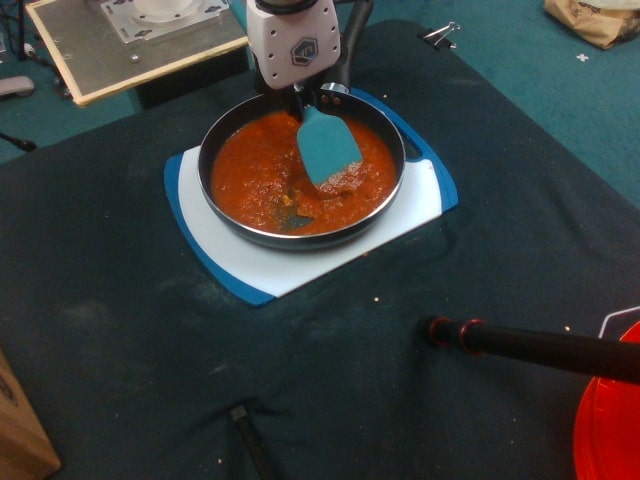}}\\

  \caption{\textbf{Visualization of 4 different cooking skills: cutting a bell pepper, peeling a carrot, scraping a cutting board, and stirring a pan.} Each of these skills was performed by a template selected by our best-performing approach, combining both language and learned optic flow features from video.}
  \label{fig:vis} 
   \squeezeup
\end{figure*}

\section{Related Work}
\subsection{Learning to Select Templates}
Behavior templates and other representations of primitive actions are commonly used in robotics to abstract away low-level control in learning and task planning. Multiple prior works have shown that learning with templates can greatly increase the efficiency of robot learning over learning in joint-torque space or learning position trajectories without sacrificing generality \cite{dalalAcceleratingRoboticReinforcement2021} \cite{nasirianyAugmentingReinforcementLearning2022}. Both these works learn a neural network to select a template to execute and any relevant parameters for that template. Schaldenbrand et al.'s work learns to paint with brushstroke templates by executing a set of templates, capturing a visual representation of them, and then using that representation to optimize a sequence of brushstrokes to complete a painting\cite{schaldenbrandFRIDACollaborativeRobot2022}. We draw inspiration from this approach to inform our method for learning from human videos by executing templates on the robot, collecting videos of the executions, and comparing them to retrieved videos of humans performing relevant tasks.

\subsection{Template Selection from Internet Data}
Several papers have explored large language models to select among primitive robot behaviors. Works like \cite{ahnCanNotSay2022a}, \cite{liangCodePoliciesLanguage2022}, and \cite{wuTidyBotPersonalizedRobot2023} used a large language model to construct task plans, write code to interact with templates, and learn human preferences. We draw specifically from Ahn et al. \cite{ahnCanNotSay2022a} to enable LLMs to select among a set of language descriptions for templates. Works including \cite{driessPaLMEEmbodiedMultimodal2023} and \cite{vergheseUserloopEvaluationMultimodal2024} have used multi-modal language models that can consider both language and visual modalities like images and videos to select templates for task planning. Finally, various methods have looked at learning or optimizing manipulation skills directly from human video. For example, Bahl et al. \cite{bahlHumanRobotImitationWild2022} used a comparison between human and robot video along with a gradient-free optimizer to learn a set of actions to manipulate articulated joints like drawers. Bahety et al. \cite{bahetyScrewMimicBimanualImitation2024} and Bahl et al. \cite{bahlAffordancesHumanVideos2023} used videos of human demonstrations to identify parameters for structured object interactions. 


\subsection{Visual Representations for Learning from Internet Data}
The right visual representation is crucial to enable robots to learn from internet images and videos. Several projects have trained vision encoders with internet datasets to accelerate robot learning \cite{nairR3MUniversalVisual2022c}\cite{radosavovicRealWorldRobotLearning2022a}\cite{majumdarWhereAreWe2024}. \cite{nairR3MUniversalVisual2022c} in particular uses image language pairs to learn a useful visual representation for downstream learning. Multiple works seeking to build video foundational models to learn representations useful for classification, retrieval, and other downstream tasks have opted to go this route by training on video-text pairs with contrastive or classification-based loss functions \cite{zhaoLearningVideoRepresentations2022}\cite{wangInternVideoGeneralVideo2022}\cite{monfortMultiMomentsTimeLearning2021}. In our work, we select LaViLa \cite{zhaoLearningVideoRepresentations2022} as the best representative example of this approach, as it is trained on egocentric video data and fine-tuned on Epic Kitchens~\cite{damenScalingEgocentricVision2018}\cite{damenRescalingEgocentricVision2020}, the dataset we use. Prior works have used these types of models to compare videos of robot behavior, and we would like to determine if this video representation is effective for our domain. Finally, optic flow is another way to represent video. Prior works have learned to predict optic flow and used extracted optic flow in training to improve robot learning approaches \cite{goyalIFORIterativeFlow2022}\cite{linFlowRetrievalFlowGuidedData2024}. Optic flow does an excellent job of encoding low-level motion between video frames. We seek to explore optic flow as an alternative representation for performing video comparisons between human demonstrations and robot behavior executions.

\section{Methods}

This work seeks to determine the efficacy of natural language and human video for selecting among a set of behavior templates to perform tool-use skills. In our setting, the robot receives a natural language label $l_s$ that specifies the skill to accomplish, the tool to use, and the objects to act on. The robot has a set of behavior templates $\Pi$ consisting of hybrid position force controllers with parameterized trajectories and force levels. Each behavior template $\pi$ has a brief natural language descriptor $l_\pi$ to describe its parameters. The robot must select a single behavior template $\pi$ to best accomplish the skill described by $l_s$.

\subsection{Template Library}
The robot is given a set of behavior templates consisting of parameterized object-centric hybrid position-force controllers. These controllers seek to apply force from a grasped tool to a recipient object while following a trajectory in the plane perpendicular to the force direction.
Our set of 33 templates is generated by combining a set of 11 basic trajectories with three different levels of applied force. The set of trajectories consists of both periodic motions like circular, forward and back, or side-to-side motions and directional motions like pushing or pulling along a direction. We use an open-vocabulary object detector \cite{zhouDetectingTwentythousandClasses2022} to identify the recipient object location and dimensions from an overhead RGB-D camera. This information is used to orient forces and motion, center the trajectories on the recipient object, and scale the trajectories by the object size. 
\begin{figure*}[t]
    \centering
    \includegraphics[width=1\columnwidth,trim={6cm 0cm 6cm 1cm}]{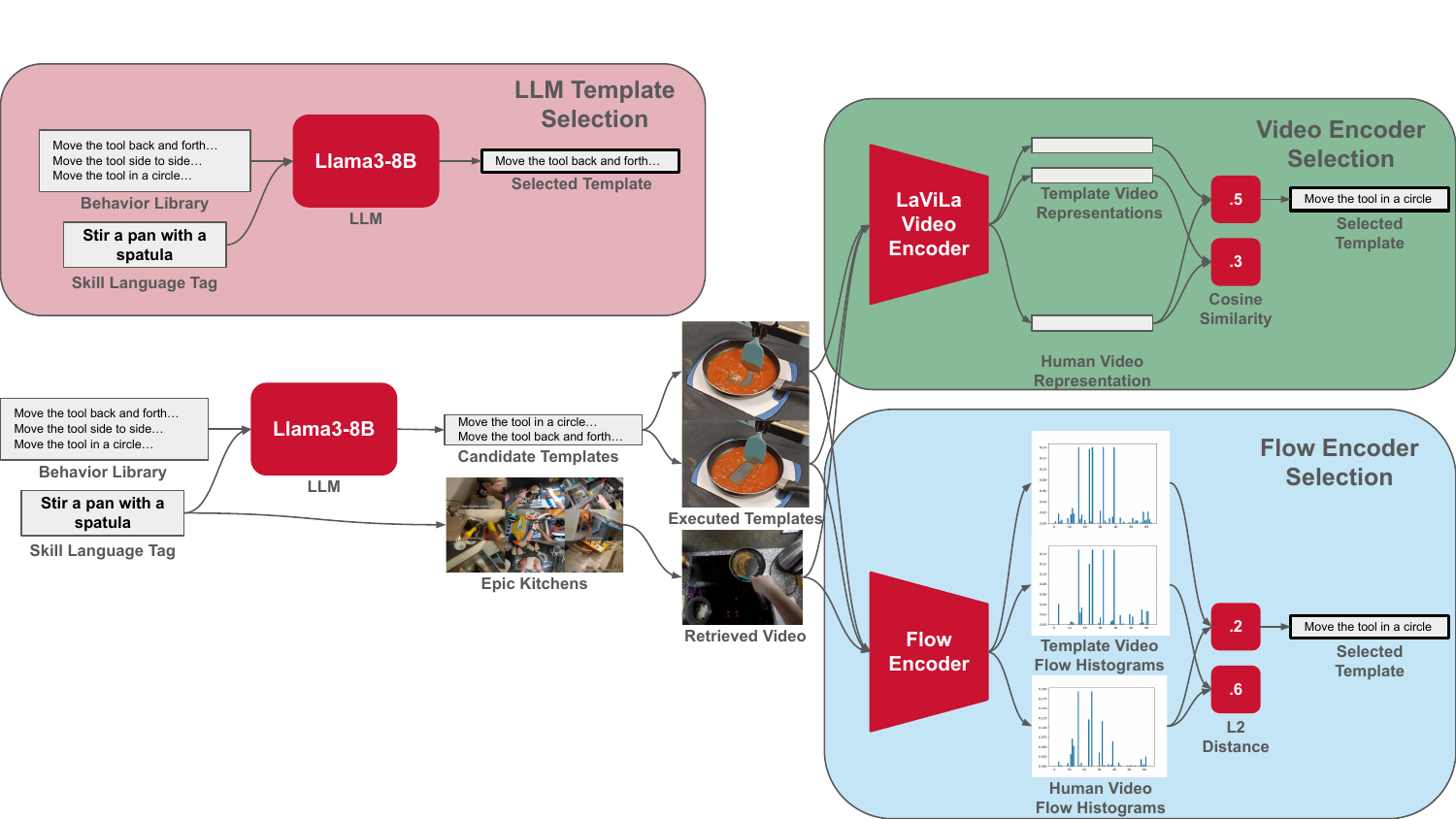}
    \caption{\textbf{Overview of three different approaches for using internet data and models to select templates.} Template selection via LLM is shown in red. The LLM is also used to select a set of candidate templates, which are then executed and the video of that execution is compared to retrieved videos of of humans performing the skill using pretrained video encoders (green) or a learned optic flow encoder (blue).}    
    \label{fig:overview}
    \squeezeup
\end{figure*}

\subsection{Template Selection with Large Language Models}
For a Large Language Model (LLM) to select among templates, the templates need some form of natural language description for the LLM to consider. We generate brief language descriptions for each template automatically based on the trajectory and force parameters it uses. While an LLM may be able to reason about the applicability of a general template for a specific skill, we find the LLM performs better if the template is described in the context of the tool and other objects involved in the skill. To accomplish this, the template language descriptions have placeholders that can be filled in. For example, the language label for the template with medium force and a small circle trajectory is:

\begin{lstlisting}
Move the [tool] in a [small circle] while applying [medium] pressure to
the [recipient]
\end{lstlisting}

where [small circle] and [medium] are the trajectory and force parameters, [tool] refers to the object grasped by the robot, and [recipient] refers to the object being acted upon. We refer to our set of templates as $\Pi$, an individual template as $\pi$, the set of template descriptors as $l_\Pi$, and an individual template descriptor as $l_\pi$

Given a set of templates and their language descriptors, as well as the natural language label describing the desired skill, the LLM outputs a score for each template, indicating the likelihood this template may be appropriate to accomplish the desired skill $p(l_\pi | l_s)$. We follow the language model scoring method outlined in \cite{ahnCanNotSay2022a}.

LLMs model the distribution of subsequent tokens conditioned on all previous tokens $p_\theta(t_{k+1} | t_k, \ldots, t_0)$. For a template language descriptor $l_\pi$ tokenized as $[t_{k+1}, \ldots, t_{k+n}]$, we can evaluate the likelihood an LLM would generate that language descriptor given a prompt as:
\begin{equation}
    p_\theta(t_{k+1}, \ldots, t_{k+n} | t_k, \ldots, t_0) = \prod_{i = k+1}^{k+n} p_\theta(t_i | t_{i-1}, \ldots, t_0) 
\end{equation}
where $[t_k, \ldots, t_0]$ is the tokenized representation of the prompt. With the prompt: 
\begin{lstlisting}
To successfully [skill] you should 
\end{lstlisting}
we evaluate the likelihood, according to the LLM, that each template might complete the given skill. In practice, we normalize the individual token probabilities in a logarithmic space to avoid penalizing templates with longer descriptors. We use the Llama3 8 billion parameter model running in 16-bit precision for all LLM evaluations.


\subsection{Template Selection by Comparison to Retrieved Video}
We can also select templates to accomplish a skill by executing those templates on the robot, collecting video of this execution, and comparing that video to videos of humans performing the same skill. Comparison to human video can offer more discriminative template selection at the cost of requiring execution of templates in the real world or a high-fidelity simulator. Due to the limited availability of simulators that can simulate complex interactions like cutting and peeling vegetables and spreading sauce, we chose to perform template execution in the real world.


\subsubsection{Retrieving Relevant Human Videos}
To compare a video of template execution on the robot to human video, we first retrieve relevant human videos from a video dataset using the skill natural language label $l_s$. In this work, we use the Epic Kitchens video dataset~\cite{damenScalingEgocentricVision2018}\cite{damenRescalingEgocentricVision2020}, which features egocentric (cameras mounted on the humans) videos segmented into semantic actions. While the Epic Kitchens dataset contains verb-noun pairs to describe each action, we find that pretrained video-language dual encoders trained with contrastive learning on paired video-text datasets perform better for retrieving relevant videos than using labels from the dataset. Using these encoders for retrieval allows our method to handle unlabeled datasets as well. To retrieve relevant videos, we encoder both the query text $l_s$ and all human videos in the shared encoder latent space and select a set of videos with the highest cosine similarity to the query text. To ensure the retrieved videos are relevant to the desired skill, we use an open-set object detector to verify the objects in the natural language skill label are present in the video. We use LaViLa~\cite{zhaoLearningVideoRepresentations2022} pretrained on Ego4D and fine-tuned on Epic Kitchens as our dual encoder for video retrieval and Detic~\cite{zhouDetectingTwentythousandClasses2022} as our open set object detector.

\subsubsection{Computing Video Similarity Scores}
The performance of human video comparison for robot template selection depends strongly on the feature space where the comparison is performed. Prior works have used large-pretrained video encoders to generate vector embeddings of videos. These encoders benefit from being trained on large amounts of data from relevant domains. However, while these video encoders do an excellent job of capturing high-level information from videos, they often lose low-level motion information, which is crucial to determining if a skill has been successfully executed. We also explore an approach to learn optic flow features as an alternative feature space.

Dense optic flow inherently represents motion between frames of a video. Furthermore, optic flow can track both how a deformable tool like a spatula moves and its effect on rigid and deformable objects in the environment. However, it is not useful to compare raw optic flow between videos. The relevant objects are not necessarily in the same position, orientation, or scale in a frame, nor are the videos temporally aligned. Instead, we want to compute an invariant set of features across both position and time to measure video similarity. We train a Vector-Quantized Variational Autoencoder (VQ-VAE) on a set of $\sim$100,000 flow frames extracted from relevant videos in the Epic-Kitchens dataset. The VQ-VAE encodes an image through a bottleneck like a normal Variation Autoencoder, but at the bottleneck, it maps each spatial feature to the nearest vector from a discrete set of vectors called a codebook. This codebook is also optimized during training to minimize reconstruction loss. Our learned flow encoder gives us a set of discrete flow features for each optical flow frame. We collect these features across a video and add them to a histogram where each bin corresponds to the number of times that vector from the codebook appeared in the flow encoding. This approach is inspired by the "bag of words" approach in natural language processing.
To determine the similarity between two videos, we calculate the L2 distance between the normalized flow feature histograms for each video. This metric evaluates the similarity between the distributions of flow features in each video, irrespective of when or where those features occur.

We use the LaViLa video encoder \cite{zhaoLearningVideoRepresentations2022} as the pretrained encoder to generate features to measure video similarity. The LaViLa encoder is trained on Ego4D, a large egocentric video dataset, and fine-tuned on Epic Kitchens, making it a reasonable video encoder for our cooking skills. To select templates with the LaViLa encoder, we score templates by the average cosine distance between template execution video, and the set of retrieved human videos. 

Optic flow is extracted using GMFLow trained on the FlyingThings3D dataset~\cite{xuGMFlowLearningOptical2022}. For our VQ-VAE flow encoder, we use the default parameters from the original paper \cite{oordNeuralDiscreteRepresentation2018} except we reduce the codebook size from 512 to 64. This forces the network to learn a compressed set of discrete feature vectors and significantly improves performance. The selected template has the smallest average L2 distance between its histogram and the histograms of all retrieved human videos.

Instead of executing all 33 templates on each of the 16 skills to provide videos for template selection, we use the LLM scores as a prior and only test the five most likely templates according to the LLM. Figure \ref{fig:overview} shows our pipeline for performing template selection with language models, the LaViLa video encoder, and our optic flow encoder.


\section{Experimental Results}
\label{sec:result}
\begin{table*}
\centering
\setlength{\tabcolsep}{3pt}
\resizebox{.99\textwidth}{!}{%
\begin{tabular}{lrrrrrrrrl}
\hline
\textbf{Method} & \textbf{Cut}  & \textbf{Peel} & \textbf{Scrape} & \textbf{Scrub} & \textbf{Slice} & \textbf{Spread} & \textbf{Stir} & \textbf{Wipe} & \textbf{Average} \\ \hline
LLM             & \textbf{0.4}            & \textbf{1.0}  & 0.0             & \textbf{0.8}   & \textbf{0.8}   & \textbf{1.0}    & 0.5           & 0.8           & 0.67             \\
Vision Encoder  & 0.2           & 0.0           & 0.5             & \textbf{0.8}   & 0.0            & \textbf{1.0}    & 0.65          & 0.85          & 0.50             \\
Flow Encoder    & 0.03          & 0.55          & \textbf{1.0}    & 0.7            & \textbf{0.8}   & \textbf{1.0}    & \textbf{0.95} & 0.9           & 0.74             \\
LLM + Flow     & 0.35 & \textbf{1.0}  & \textbf{1.0}    & 0.7            & \textbf{0.8}   & \textbf{1.0}    & 0.5           & \textbf{0.95} & \textbf{0.79}    \\ \hline
\end{tabular}
}
\caption{\textbf{Success rate from independent human evaluators for each type of skill.} Human evaluators were asked to determine whether a video of a selected template executing a skill was successful. The average success rate across ten human evaluations is reported. }
\label{tab:success}
\squeezeup
\end{table*}
 \begin{figure*}[!t] 
    \centering
  \subfloat{%
       \includegraphics[width=0.5\linewidth]{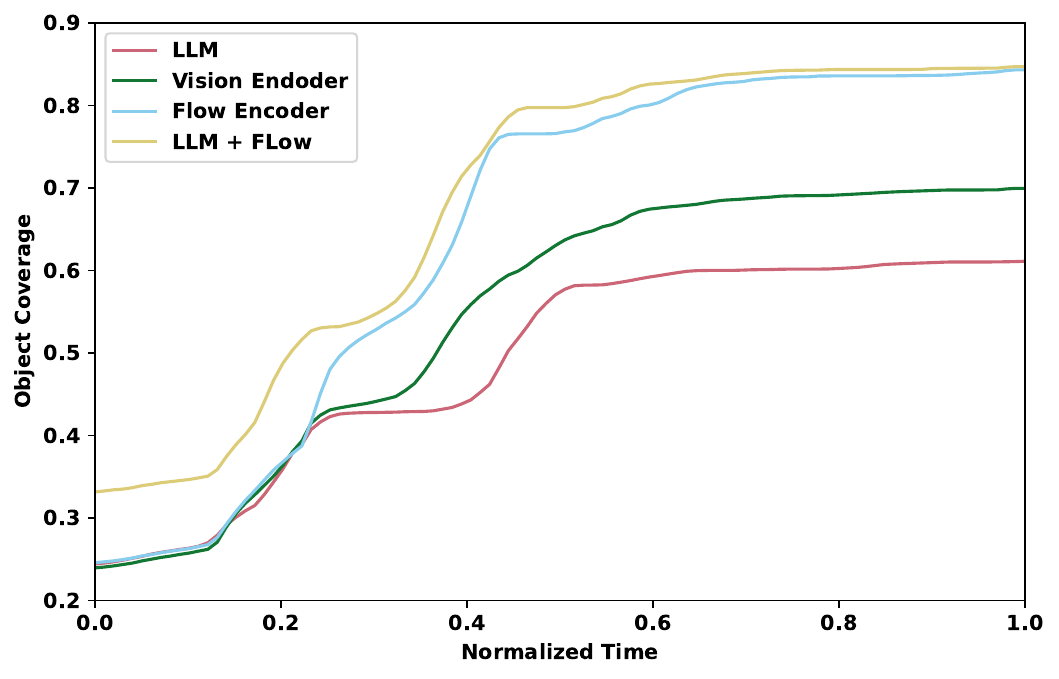}}
  \subfloat{%
        \includegraphics[width=0.5\linewidth]{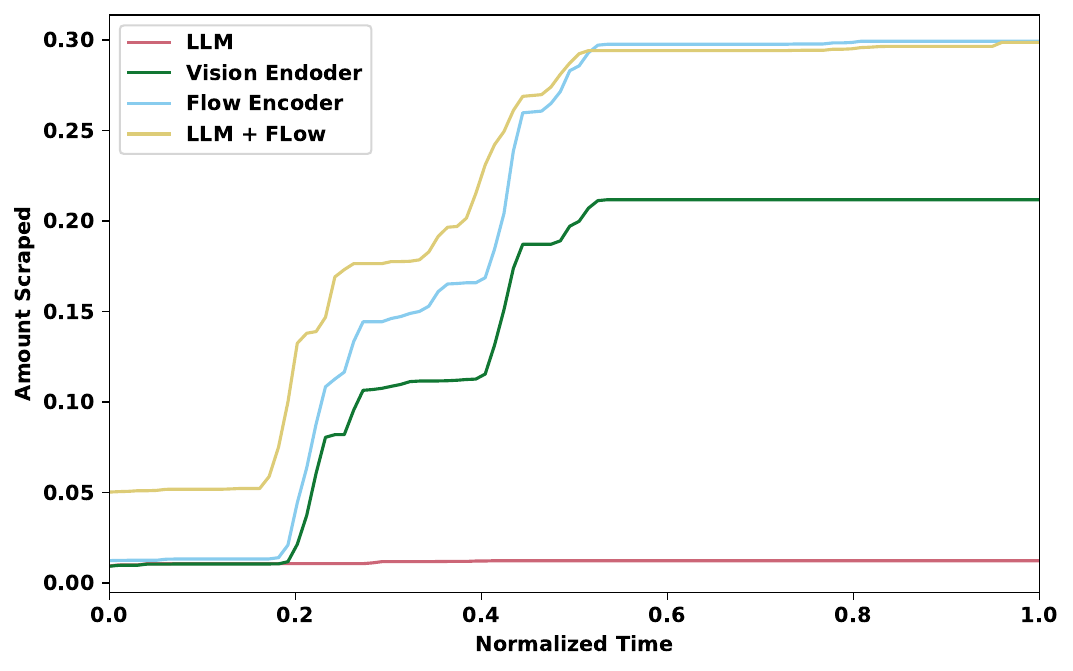}}

  \caption{\textbf{Progress of the wiping (left) and scraping (right) skill execution for various template selection methods.} The X-axis is the normalized episode length. The Y-axis is task progress. Task progress is determined for wiping as the percentage of the recipient object that the cloth has wiped in pixel space. Task progress is determined for scraping as the amount of orange pepper at the edge of the cutting board in pixel space.}
  \label{fig:progress} 
   \squeezeup
\end{figure*}
\subsection{Experimental Setup}

Templates are performed by an xArm7 robot equipped with a 6-axis force-torque sensor and parallel jaw gripper. For template execution, the robot starts with the tool object grasped. Functional grasps for tool use is a well-studied problem and is outside the scope of this work. During execution, the recipient object is fixed to the table to prevent movement. This is similar to the case where a second arm might hold the object in place. Video of the template executions is recorded by an overhead RealSense D435 RGB-D camera. This camera is also used to localize the recipient object using an open-vocabulary object detector (Detic~\cite{zhouDetectingTwentythousandClasses2022}) and identify its dimensions.

We test the selections from the LLM, pretrained video encoder, flow encoder, and a weighted combination of the LLM and flow encoder scores: $S_{combined} = \lambda S_{LLM} + (1 - S_{flow})$, where $S_{LLM}$ are the normalized likelihoods from the LLM selector and $S_{flow}$ are the normalized histogram distances from the flow encoder. These selections are evaluated across 16 skills comprising four cutting skills (cutting carrots, cucumbers, bell peppers, and mushrooms), two peeling skills (peeling carrots and cucumbers), two stirring skills (stirring chopped peppers and stirring tomato sauce), two scraping skills (scraping a cutting board with a knife, and with a bench scraper), two scrubbing skills (scrubbing a cutting board and a plate), two wiping skills (wiping a cutting board and a plate), spreading tomato sauce, and slicing a pizza.

The final templates selected by each method are re-evaluated on a variation of the skill. These variations may include different versions of tools, different sizes or colors of vegetables, and different consistencies of sauces. We evaluate the success and quality of select skills using computer vision techniques and all skills using human evaluators. Evaluators are asked to view all the videos for a given skill, and indicate for each video if the skill was performed successfully and give a quality rating from 1-5 for the skill (normalized to 0 to 1 for visualization), with 1 indicating poor performance and 5 indicating excellent performance. The quality ratings are given independent of the success rating and give us an additional performance signal in cases where success may be very easy or difficult. Ratings were collected from 10 people. Videos of template execution on the skill variants used for evaluation can be found at \url{https://sites.google.com/view/skillsmadetoorder/home}

\subsection{Results}
\begin{figure*}[t]
    \centering
    \includegraphics[width=1\textwidth]{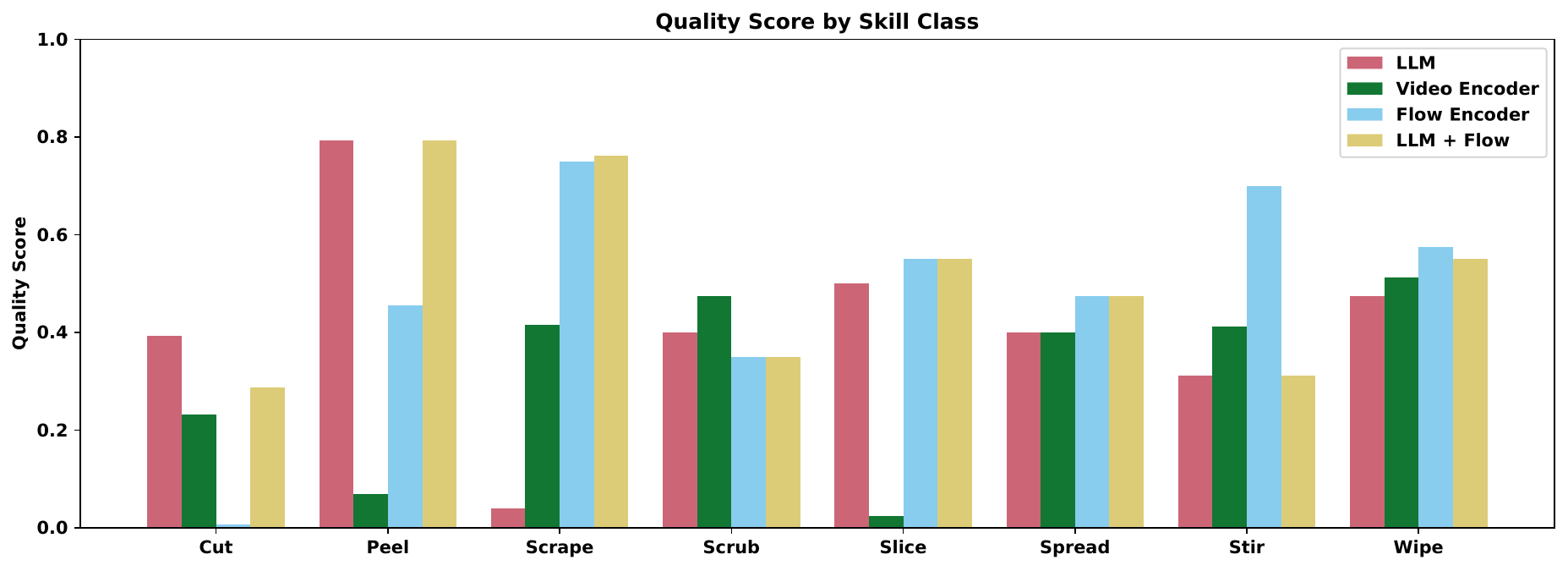}
    \caption{\textbf{Quality scores from human evaluators for each type of skill}. Human evaluators were asked to rate how well the template chosen by each approach performed the skill. The approaches tested were an LLM-based template selector, template selection by comparing executed templates to human video using features from a pretrained video encoder (LaViLa), the same comparison using learned optic flow features, and a combination of LLM and flow.}    
    \label{fig:quality}
    \squeezeup
\end{figure*}
\subsubsection{Large Language Models Can Perform Template Selection Without Visual Information}

Large language models do not process any visual input when performing template selection. While this makes them very inexpensive to run across a large library of templates, they cannot condition their selection on images or video, making it more difficult to account for details of the task setup or how well candidate behaviors work in general. Despite this limitation, the LLM in our study was able to select successful templates two-thirds of the time (table \ref{tab:success}). Furthermore, the optic flow-based encoder outperformed the LLM while only selecting among the LLM's top 5 templates instead of the whole library. This result suggests that even in cases where the LLM does not directly pick a good template for a skill, a good template may exist in the LLM's top choices. In conjunction with their ability to quickly score an entire template library, this result suggests that LLMs have great potential as a filter for template selection, picking high-likelihood templates to pass to other expensive but higher-quality selectors.

\subsubsection{Learned Optic Flow Encoding Outperforms Pretrained Video Encoders}

Table \ref{tab:success} shows that our learned flow encoder significantly outperforms the pretrained LaViLa video encoder across almost every type of skill. This result occurs despite the LaViLa video encoder being trained on significantly more video frames from Ego4D and EpicKitchens (pretrained on 4 million video-text pairs from Ego4D and fine-tuned on multiple samples from 67 thousand video clips in Epic Kitchens~\cite{zhaoLearningVideoRepresentations2022}). Dense optic flow contains rich information about motion but relatively little information about objects and their appearance. Conversely, while video encoders do get some information about motion between video frames, because they subsample frames from the video (LaViLa subsamples 16 evenly spaced frames), they lose a lot of low-level motion information. This result indicates that when comparing videos for the purposes of template selection or other scoring tasks, this low-level motion information is more significant than high-level semantic features.

For example, figure \ref{fig:progress} shows the portion of the object that has been wiped, and the amount of food scraped as a function of time for each of our approaches. In the wiping skill, the features from the optic flow encoder select a side-to-side wiping technique that was present in human video over the sub-optimal back-and-forth wiping technique that the video encoder selected. In the scraping skill, the flow encoder reliably selected templates with medium force and a side-to-side trajectory to clear food from the cutting board. On the other hand, the vision encoder selected a template with a back-and-forth trajectory when using the bench scraper that failed to effectively clear food from the cutting board. These plots also highlight the advantages of selecting templates using video comparison, as both video comparison methods outperform the templates selected by the LLM.

\subsubsection{Good Synergies Exist Between Language and Vision Internet Data}
Examining table \ref{tab:success} shows that while encoding optic flow for selection generally outperforms language-based selection, this performance gap is inconsistent, and each modality has its own strengths and weaknesses. Language-based template selection performs better for skills like cutting and peeling, while vision-based selection performs better for scraping, stirring, and wiping skills. This is partly due to the varying extents of visual change during the execution of these two groups of skills. When cutting or peeling vegetables, the visual change in the scene outside of the motion of the tool itself is minimal. However, during scraping, stirring, and wiping, chopped vegetables are pushed around, sauce flows around a spatula, and a cloth is dragged along an object. These all induce significant visual change and optic flow in the scene. This contrast in visual change can be seen in figure \ref{fig:vis}. Conversely, as our large language model processes no visual input, it can struggle when the optimal template depends on the scene. For example, cutting is almost always performed with the blade pointing away from the user and moving along the same axis that the blade is pointing. However, many stirring methods exist, and the best method depends entirely on the items in the pan that need to be stirred.

These varying strengths create a nice synergy between the two modalities of internet data. This point becomes especially apparent when we find that linearly combining the two template selectors leads to even better performance than each of them individually. We found the optimal value of weighting on the LLM final selection ($\lambda$) to be $.1$. Recall the LLM is also acting as a filter to determine which behaviors are to be rated by the optic flow encoder method. In addition, figure \ref{fig:quality} shows that this combination of synergistic selectors had the most consistent performance quality across all types of skills. According to human evaluators, this combination of template selectors achieves a given skill 79\% of the time. This performance also comes despite our limiting assumption that a skill is achievable by a single template.



\section{Conclusion}
\label{sec:conclusion}
In this study, we explored the relative abilities of various forms of internet data to select templates to complete robot cooking skills. We sought to answer two questions: which of these forms of internet data perform the best, and do synergies exist between forms of data? We found that LLMs are capable template selectors and excellent filters for video comparison methods. We also found that discrete optic flow features are much more effective for comparing videos of template execution to reference videos of humans compared to pretrained video encoders. This result has interesting implications for learning reward models from videos that we would like to explore further in our work. Importantly, we discovered significant synergies between language-based template selection and template selection via comparison to human video. Trivially combining these two methods gave us our best-performing approach, yielding a 79\% success rate across 16 cooking skills.

\subsection{Future Work}
Recently, multi-modal LLMs that process images or video have become increasingly capable and accessible. In future work, we would like to explore the ability of these multimodal LLMs to improve text-based template selection by conditioning on a visual representation of the environment. In this work, the robot had to execute a set of candidate templates for every skill to select via comparison to human video. We believe it is possible to cache videos of prior template executions and use them to select templates for related skills. This capability could be further augmented by leveraging generative image and video models to replace relevant objects and tools from cached videos with.
Finally, rather than hand designing a set of trajectories for templates, we would also like to explore extracting these trajectories from human video. Overall, we believe this study presents a different look into methods for learning contact-rich skills from internet data, and we believe this direction strongly warrants further investigation.

\bibliographystyle{IEEEtran}
\bibliography{root}

\end{document}